\documentclass[a4paper,fleqn]{cas-sc}

\usepackage[authoryear]{natbib}
\usepackage{graphicx} 
\usepackage{float}
\usepackage{algorithm}  
\usepackage{algpseudocode}
\usepackage{color}
\usepackage{setspace}

\def\tsc#1{\csdef{#1}{\textsc{\lowercase{#1}}\xspace}}
\tsc{WGM}
\tsc{QE}
\tsc{EP}
\tsc{PMS}
\tsc{BEC}
\tsc{DE}

\begin{document}
\let\WriteBookmarks\relax
\def\floatpagepagefraction{1}
\def\textpagefraction{.001}
\shorttitle{Prediction of Fault Slip Tendency in CO${_2}$ Storage using Data-space Inversion}
\shortauthors{He et al.}

\title [mode = title]{Prediction of Fault Slip Tendency in CO${_2}$ Storage using \\ Data-space Inversion}

\author[1]{Xiaowen He\corref{cor1}}[orcid=0009-0006-3024-1034]
\cormark[1]
\ead{xiaowen@stanford.edu}

\author[2]{Su Jiang}[orcid=0000-0002-5744-2526]

\author[1]{Louis J.~Durlofsky\corref{}}[orcid=0000-0003-2733-6543]

\address[1]{Department of Energy Science and Engineering, Stanford University, Stanford, California, USA}
\address[2]{Department of Civil and Environmental Engineering, Carnegie Mellon University, Pittsburgh, Pennsylvania, USA}

\begin{abstract}
Accurately assessing the potential for fault slip is essential in many subsurface operations. Conventional model-based history matching methods, which entail the generation of posterior geomodels calibrated to observed data, can be challenging to apply in coupled flow-geomechanics problems with faults. In this work, we implement a variational autoencoder (VAE)-based data-space inversion (DSI) framework to predict pressure, stress and strain fields, and fault slip tendency, in $\text{CO}_2$ storage projects. The main computations required by the DSI workflow entail the simulation of ${\mathcal O}(1000)$ prior geomodels. The posterior distributions for quantities of interest are then inferred directly from prior simulation results and observed data, without the need to generate posterior geomodels. The model used here involves a synthetic 3D system with two faults. Realizations of heterogeneous permeability and porosity fields are generated using geostatistical software, and uncertain geomechanical and fault parameters are sampled for each realization from prior distributions. Coupled flow-geomechanics simulations for these geomodels are conducted using GEOS. A VAE with stacked convolutional long short-term memory layers is trained, using the prior simulation results, to represent pressure, strain, effective normal stress and shear stress fields in terms of latent variables. The VAE parameterization is used with DSI for posterior predictions, with monitoring wells providing observed pressure and strain data. Posterior results for synthetic true models demonstrate that the DSI-VAE framework gives accurate predictions for pressure, strain, and stress fields and for fault slip tendency. The framework is also shown to reduce uncertainty in key geomechanical and fault parameters.
\end{abstract}



\maketitle 


\doublespacing

\section{Introduction}
Faults may be present in many of the formations targeted for geological carbon storage. The megatonne-scale injection of $\rm{CO}_2$ can potentially destabilize these faults by altering pressure and stress conditions, resulting in fault slip and induced seismicity. Thus, an understanding of the stress-state evolution of pre-existing faults is essential for the safe management of storage operations. Predicting fault slip and induced seismicity can be challenging, however, due to the significant uncertainties in the geomechanical and flow properties of subsurface systems. Many studies have applied model-based history matching, in which geomodel properties are calibrated such that the resulting simulation predictions match observations. In the case of coupled flow-geomechanics in faulted systems, model-based inversion procedures can be extremely costly given the time-consuming nature of these simulations and the large number of forward runs required.

Our goal in this work is to apply an alternate approach, namely data-space inversion (DSI), to directly predict stress fields, fault-slip behavior, and other important quantities in carbon storage settings. In DSI procedures, rather than generate posterior (history matched) geomodels, predictions for quantities of interest (QoI), conditioned to observed data, are constructed directly. This approach requires a substantial number, e.g., $\mathcal{O}(1000)$, of prior flow simulations, though no additional simulations are required during history matching. Thus, the impact of different amounts and types of data, different levels of data precision, etc., can be quickly assessed. The key limitation of DSI methods is that posterior models are not available for subsequent applications such as well placement optimization.

Fault reactivation can occur in many subsurface activities, including wastewater disposal \citep{vadacca2021slip}, geothermal energy production \citep{gan2014analysis}, and CO$_2$ storage operations \citep{zakharova2014situ}. In geological carbon storage (GCS) settings, fault reactivation can lead to seismicity or to CO$_2$ leakage. At the In Salah (Algeria) project \citep{shi2012}, for example, leakage into the overlying formation was observed. In the Quest CCS project in Alberta, Canada \citep{bourne2014_Quest, sarkarfarshi2019_Quest} and at the Decatur CCS site in Illinois, USA \citep{kaven2015_Decatur}, fault reactivation due to overpressurization occurred. The risk associated with fault reactivation can be quantified in terms of fault slip tendency, or FST \citep{morris1996slip}. FST is defined as the ratio of shear stress to effective normal stress acting on a fault. This quantity cannot be measured directly, so coupled numerical simulations are used to predict the state of stress and thus FST \citep{khan2024geomechanical, rutqvist2017overview, song2023geomechanical}. The inherent uncertainty in flow and geomechanical rock properties leads to a wide range of possible outcomes. This prior uncertainty can be reduced through application of data assimilation, which we now discuss. 

Model-based data assimilation methods, using various types of data, are typically applied to calibrate geological models and reduce prediction uncertainty.
\citet{seabra2024ai}, for example, applied pressure monitoring data to calibrate permeability parameters and predict pressure fields in GCS. \citet{han2025accelerated} integrated monitoring well data (pressure and $\rm{CO_2}$ saturation) and surface displacement measurements to infer metaparameters and geological realizations using a hierarchical Markov chain Monte Carlo (MCMC) method. \citet{wang2025deep} assimilated 4D seismic saturation fields and monitoring well data to calibrate geomodel parameters and predict saturation fields. \citet{liu2024geostatistical} applied seismic data to infer rock properties including porosity and clay volume. \citet{anyosa2024evaluating} integrated seismic and electromagnetic monitoring data to reduce uncertainty in CO$_2$ plume predictions and proposed an efficient sequential monitoring strategy. While strain measurements provide valuable insights into geomechanical effects \citep{guglielmi2020complexity}, few studies have explored data assimilation using both strain and pressure data to predict fault slip in coupled flow-geomechanics systems. This may be due, in part, to the high costs associated with the computations required for this application.

To accelerate history matching computations, deep learning-based surrogate models are now being employed.
Along these lines, \citet{tang2020} designed a recurrent R-U-Net surrogate model that combines a residual U-Net and a long short-term memory recurrent network to predict pressure, saturation, and well rates in oil-water systems. This architecture was extended to treat coupled flow-geomechanics problems \citep{tang2021deep, han2025accelerated}. \citet{wen2022} presented a U-FNO model based on Fourier neural operators to predict $\rm{CO}_2$ saturation and pressure fields.
\citet{ju2024learning} proposed a graph network-based surrogate model to provide pressure and CO$_2$ saturation in geologically complicated faulted systems. \citet{tang2025} also developed a graph network-based procedure applicable for varying horizontal well configurations in GCS problems.

Even with advanced surrogate models, model-based inversion methods can still face challenges. This can occur when the underlying system is characterized by a high degree of prior uncertainty, when uncertainty cannot be represented systematically in terms of a tractable set of searchable parameters, or when surrogate model training (which includes performing the training simulation runs) is itself highly challenging or time consuming. 
As an alternative to traditional approaches, within the context of oil reservoir simulation, \citet{sun2017a} introduced the DSI procedure. Unlike model-based methods, DSI focuses on the prediction of particular QoI (flow rates at wells, states at specific locations and times), which leads to enhanced robustness and computational savings. DSI has since been extended and applied within oil field settings by, e.g., \citet{lima2020data}, \citet{jiang2021treatment} and \citet{hui2023data}, and within enhanced geothermal settings by \citet{de2025data}. \citet{sun2019} applied DSI for GCS problems, with $\rm{CO}_2$ saturation in the top layer of the model predicted using measurements from monitoring wells. Posterior sampling in DSI procedures is best accomplished when the key quantities are parameterized in terms of Gaussian distributed latent variables. \citet{jiang2024} introduced an adversarial autoencoder (AAE) with 3D convolutional long short-term memory (ConvLSTM) layers to parameterize pressure and $\rm{CO}_2$ saturation fields. The 3D ConvLSTM layers effectively captured high-dimensional spatio-temporal data, while the AAE ensured the latent variables were Gaussian distributed. Although DSI does not provide posterior models, it was recently used by \citet{barros2025fast} in a closed-loop framework involving the optimization of future well locations.

While the DSI framework has been widely used for flow-only problems, its potential in flow-geomechanical contexts has not been explored. In this work, we address this gap by introducing a DSI framework able to predict stress states and fault slip tendency in problems involving coupled flow and geomechanics. A variational autoencoder (VAE) is used to parameterize the coupled flow-geomechanics responses. This VAE directly imposes a probabilistic structure on the latent space through the variational loss function, making it easier to train than an AAE. 
The VAE-based DSI framework is applied to predict coupled flow-geomechanics responses in faulted systems. The prior simulation data required by DSI are generated by performing coupled simulations on multiple realizations of 3D faulted geomodels using the open-source multiphysics simulator GEOS~\citep{bui2021}. These simulation results also provide the training data for the VAE. The geomodel and simulation setup are partially based on the case introduced by \citet{silva2023}, which is representative of a Gulf of Mexico storage project. By assimilating in-situ pressure and strain observations, the extended DSI-VAE method is shown to provide accurate posterior pressure, strain, and stress fields, along with reduced uncertainty in fault slip tendency. Key geomechanical and fault parameters are also estimated.

This paper proceeds as follows. In Section~\ref{Sec: Geomodel and DSI Methodology}, we first describe the 3D faulted geomodel developed in this work. The detailed permeability field and key geomechanical and fault parameters are taken to be uncertain. Next, the overall DSI-VAE framework for coupled flow-geomechanics systems is introduced. The VAE used for data parameterization is then briefly described (details are provided in online Supporting Information, SI). The VAE is trained using coupled flow-geomechanics simulation results for 1200 geomodels sampled from the prior. 
In Section~\ref {Sec: VAE reconstruction results}, we demonstrate VAE performance through various assessments. Detailed DSI results for synthetic true models are presented in Section~\ref{Sec:DSI_results}. \textcolor{black}{Timing considerations are discussed in Section~\ref{sec:costs}.} A summary and suggestions for future research are provided in Section~\ref{Sec: Conclusions}. Details on the simulation models and the VAE, a description of the optimization procedure used to determine monitoring well locations, and history matching results for other synthetic true models are provided in SI.

\section{Geomodel and DSI methodology}\label{Sec: Geomodel and DSI Methodology}
In this section, we present the specially designed geomodel and the data-space inversion (DSI) framework applied in this work. We introduce a variational autoencoder (VAE) for parameterizing the spatio-temporal pressure, strain, and stress fields. The VAE latent variables are then utilized in DSI. 

\subsection{Faulted geomodel}\label{Subsec: Faulted geomodel}
The 3D faulted geomodel considered in this study is partially derived from a realistic Gulf of Mexico model developed by \citet{silva2023}. Aspects of the model are modified because our interest here is in predicting stresses for cases with fault slip tendency near the fault friction coefficient.
Fault slip tendency ($T_\mathrm{s}$) represents the ratio of shear stress magnitude ($\tau$) to effective normal stress magnitude ($\sigma_n^\prime$) along the fault \citep{morris1996slip}, i.e.,
\begin{equation}
    T_\mathrm{s}=\frac{\tau}{\sigma_n^\prime}.
\label{Eq: Ts = tau/sigma_n'}
\end{equation}
To compute the effective normal stress and shear stress, we first calculate the traction vector $\bf{t} = \boldsymbol{\sigma} \cdot \bf{n}$ along the fault, where ${\boldsymbol \sigma}$ is the stress tensor and $\bf{n}$ is the unit normal vector. The magnitude of effective normal stress is then given by $\sigma_n^\prime=\sigma_n - \alpha p$, where $\sigma_n = \bf{n} \cdot \bf{t}$, $\alpha$ is the Biot coefficient, and $p$ is pore pressure. The magnitude of shear stress is computed as $\tau = \sqrt{| {\bf t} |^2-\sigma_n^2}$. In this work, consistent with previous studies~\citep{byerlee1978friction,ikari2011relation,urpi2016dynamic}, we consider the fault friction coefficient to be 0.6. This means slip can be expected to occur if $T_\mathrm{s} > 0.6$.

\textcolor{black}{The model and simulation grid are shown in Fig.~\ref{Fig: realization, injectors, monitors, & fault mesh}. The overall domain includes the central storage aquifer (shown in Fig.~\ref{Fig: realization, injectors, monitors, & fault mesh}(a) and (b) and highlighted in (c) and (d)), surrounding region (evident in Fig.~\ref{Fig: realization, injectors, monitors, & fault mesh}(c)), and caprock and basement rock (shown in Fig.~\ref{Fig: realization, injectors, monitors, & fault mesh}(d)).} The storage aquifer measures 25~km $\times$ 27~km $\times$ 60~m and is discretized using 50 $\times$ 50 $\times$ 20 cells (50,000 cells total). The full domain spans 33.5~km $\times$ 34.5~km $\times$ 2660~m, and is discretized with 60 $\times$ 60 $\times$ 35 cells (126,000 cells total). The surrounding formation, caprock, and basement are modeled at lower resolution than the storage aquifer.

\begin{figure}
    \centering
    \includegraphics[width=0.8\textwidth]{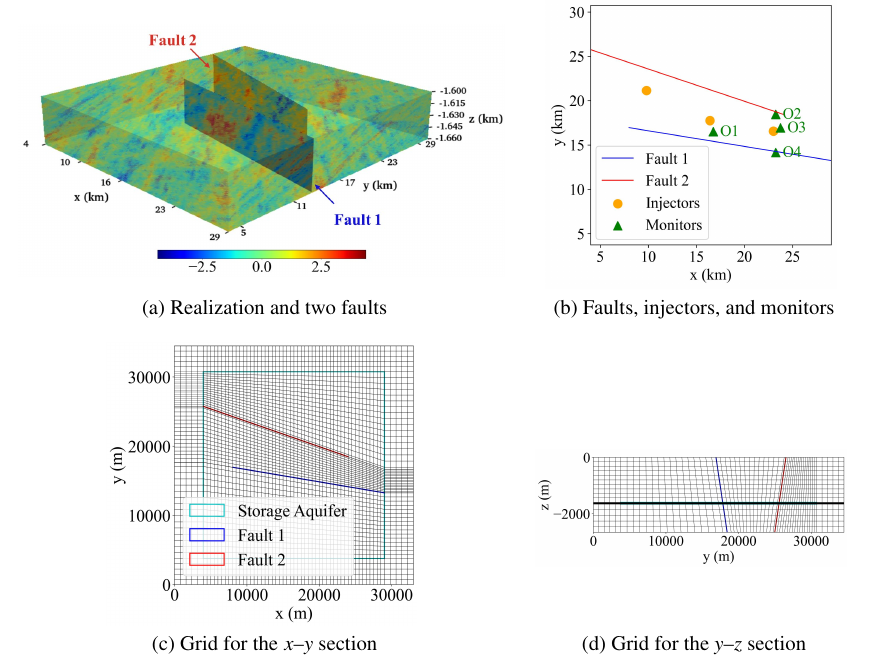}
    \caption{\textcolor{black}{Model setup and simulation grid. (a) Storage aquifer realization and two faults in the model. (b) Locations of faults, injectors, and monitoring wells in the model in the $x$--$y$ plane at a depth of 1630~m. (c) Grid in the $x$--$y$ plane at a depth of 1630~m. (d) Grid in the $y$--$z$ section at $x = 4000$~m.}}
    \label{Fig: realization, injectors, monitors, & fault mesh}
\end{figure}

We generate heterogeneous multi-Gaussian log-permeability and porosity fields for the storage aquifer. An example realization is shown in Fig.~\ref{Fig: realization, injectors, monitors, & fault mesh}(a). The variogram and scaling parameters, which include correlation lengths in the $x$, $y$, and $z$ directions, azimuth, dip angle, mean and standard deviation of log-permeability and porosity, and permeability anisotropy ratio, are provided in Table~\ref{Table: realization params & prior params}. These values are based on those used by \citet{silva2023}, with some modifications to ensure relatively large $T_\mathrm{s}$ in the resulting simulations. The detailed (random) realizations are generated using the geological modeling software SGeMS~\citep{remy2009applied} on 100 $\times$ 100 $\times$ 40 regular grids. Nearest-neighbor interpolation is then applied to map these realizations to the irregular grids in our model. For the caprock, the porosity is set to 0.08, the horizontal permeability to 0.001~mD, and the vertical permeability to 0.0001~mD \citep{walters2023geochemistry}. In the basement rock, the porosity is 0.09, the horizontal permeability is 0.23~mD, and the vertical permeability is 0.023~mD \citep{dutton2010reprint}. For the surrounding region, we set porosity to 0.08, horizontal permeability to 20~mD, and vertical permeability to 2~mD.

\begin{table}
    \centering
    \caption{Variogram and scaling parameters for storage aquifer realizations, and prior ranges for geomechanical and fault parameters.}
    \begin{tabular}{cc}
        \toprule
        \textbf{Parameter} & \textbf{Value} \\
        \midrule
        Correlation length in $x$ direction & 7500~m \\
        Correlation length in $y$ direction & 6750~m \\
        Correlation length in $z$ direction & 7.5~m \\
        Azimuth & \(45^\circ\) \\
        Dip angle & \(45^\circ\) \\
        Mean of log-permeability ($\log_e k$, $k$ in mD) & 3 \\
        Standard deviation of log-permeability & 1.5 \\
        Mean of porosity & 0.12 \\
        Standard deviation of porosity & 0.05 \\
        Permeability anisotropy ratio $\left(k_z/k_x\right)$ & 0.1 \\
        Young's modulus ($E$) & [10, 20] GPa \\
        Poisson ratio ($\nu$) & [0.25, 0.30] \\
        Biot coefficient ($\alpha$) & [0.8, 1.0] \\
        $\log_{10}$ fault permeability multipliers ($\log_{10}\left(k_\mathrm{f}^1 / k\right)$ and $\log_{10}\left(k_\mathrm{f}^2 / k\right)$) & [-3, 0) \\
        Initial stress coefficient ($\gamma$) & (0, 1)\\
        \bottomrule
    \end{tabular}
    \label{Table: realization params & prior params}
\end{table}

All geomodel realizations include the two faults shown in Fig.~\ref{Fig: realization, injectors, monitors, & fault mesh}. Fault~1 (consistently depicted in blue) is oriented at \(10^\circ\) to the $x$-axis in the $x$-$y$ plane with a dip angle of \(60^\circ\). Fault~2 (in red) is oriented at \(20^\circ\) to the $x$-axis in the $x$-$y$ plane, and has a dip angle of \(60^\circ\). Faults in geological models can be represented through various approaches, including as zero-thickness surfaces \citep{jha2014coupled_surfaces} or as finite-thickness elements \citep{figueiredo2015coupled_elements, rinaldi2014geomechanical_elements}. In our setup, we model both faults as 10-meter-thick elements. The irregular mesh, shown in Fig.~\ref{Fig: realization, injectors, monitors, & fault mesh}(c) and (d), is designed such that cells tilt gradually as they approach the faults.

The system includes three vertical wells (represented by orange circles) positioned between the two faults, as shown in Fig.~\ref{Fig: realization, injectors, monitors, & fault mesh}(b). These wells fully penetrate the storage aquifer, and each well injects 1~Mt $\text{CO}_2$ per year for 50~years (total of 3~Mt $\text{CO}_2$/year injected). The monitoring well locations, shown as green triangles (O1--O4) in Fig.~\ref{Fig: realization, injectors, monitors, & fault mesh}(b), are determined through an optimization procedure (described in SI) similar to that used by \citet{sun2019}.

We consider the geomechanical and fault parameters to be uncertain, though unlike permeability and porosity they are taken to be constant over the full domain. These parameters include Young's modulus $E$, Poisson ratio $\nu$, Biot coefficient $\alpha$, fault permeability multipliers, denoted as $\log_{10}\left(k_\mathrm{f}^1 / k\right)$ and $\log_{10}\left(k_\mathrm{f}^2 / k\right)$, and an initial stress coefficient $\gamma$ that will be described later. The prior distributions of these parameters are assumed uniform, with ranges specified in Table~\ref{Table: realization params & prior params}. The fault permeability multipliers modify the geostatistically assigned permeabilities for the cells along the faults. For Fault~1, for example, $k_i^f=k_i^1 \times (k_\mathrm{f}^1 / k)$, where $k_i^f$ denotes the modified permeability for Fault~1 cells, $k_i^1$ is the original SGeMS permeability, and $i$ denotes any cell along Fault~1. Permeability for Fault~2 is assigned analogously.

The model is assumed to be in a normal stress regime, which means vertical stress is the maximum principal stress \citep{zoback2010reservoir}. The maximum horizontal stress is taken to be in the $x$-direction, and the minimum horizontal stress is in the $y$-direction. Letting $\sigma_{zz}^\prime$, $\sigma_{xx}^\prime$, and $\sigma_{yy}^\prime$ denote the effective stress in the $z$, $x$, and $y$-directions, we have
\begin{equation}
    \sigma_{zz}^\prime > \sigma_{xx}^\prime > \sigma_{yy}^\prime.
\end{equation}
The minimum effective horizontal stress is estimated using the Poisson ratio and maximum effective vertical stress as \citep{eaton1969fracture, meng2018analysis, zoback2010reservoir},
\begin{equation}
    \sigma_{yy}^\prime = \frac{\nu}{1-\nu} \sigma_{zz}^\prime.
\end{equation}
For $\sigma_{xx}^\prime$, which lies between $\sigma_{yy}^\prime$ and $\sigma_{zz}^\prime$, we use the coefficient $\gamma \in \left(0,1\right)$ to introduce uncertainty in the initial stress state across realizations through use of
\begin{equation}
    \sigma_{xx}^\prime = \gamma \sigma_{yy}^\prime + \left(1 - \gamma\right) \sigma_{zz}^\prime.
\end{equation}

The coupled flow-geomechanics simulations are performed using the open-source simulator GEOS \citep{bui2021}. The solid mechanics problem is treated as quasi-static. We use the two-phase, two-component ($\text{CO}_2$ and brine phases/components) fluid model in GEOS. The density of supercritical $\text{CO}_2$ in the storage aquifer is about 520~$\mathrm{kg/m^3}$. Brine salinity is set to $10^5$~ppm \citep{kraemer1984occurrence,silva2023}. Additional information on simulation inputs, including relative permeability and capillary pressure curves, are provided in SI.

The model is initialized with brine saturation of 1 and hydrostatic pressure. At the center of the storage aquifer (1630~m depth), the initial pressure and temperature are 17~MPa and $80^{\circ}\text{C}$ \citep{silva2023}. No-flow boundary conditions are imposed on all boundaries of the full domain. For geomechanical boundary conditions, roller constraints (zero normal displacement) are applied on the boundaries of the full domain, except at the top surface, which remains free \citep{silva2023}. The GEOS simulation runs performed in this work each require about 2~hours using an AMD EPYC-7543 32-Core CPU.

\subsection{DSI formulation}\label{Subsec: DSI formulation}
As discussed earlier, DSI methods provide posterior predictions (within a Bayesian framework) rather than a set of history matched realizations. The DSI workflow proceeds as follows. We begin by generating an ensemble of $N_\text{r}$ prior geological realizations ($N_\text{r}=1200$ in this work), denoted by $\textbf{m}_i \in \mathbb{R}^{N_\text{b} \times 1}$, where $i = 1, 2, \ldots, N_\text{r}$, with $N_\text{b}$ denoting the number of grid blocks in the model. These models are all represented on the same mesh, described in Section~\ref{Subsec: Faulted geomodel}. The permeability and porosity fields are heterogeneous, while the geomechanical parameters (e.g., Young's modulus and Poisson ratio) are taken to be homogeneous over the full domain. 

We perform GEOS simulations (coupled flow-geomechanics) on each prior model. These prior simulation results are expressed as 
\begin{equation}
    \left(\textbf{d}_{\text{full}}\right)_i = g\left(\textbf{m}_i\right), \ \ i = 1, 2, \ldots, N_\text{r},
\end{equation}
where $\left(\textbf{d}_{\text{full}}\right)_i \in \mathbb{R}^{N_\text{full} \times 1}$ represents the simulation results for the quantities of interest for geomodel $\textbf{m}_i$ and $g$ denotes the forward model (i.e., the simulation procedure) that maps model parameters to model responses. As demonstrated by \citet{jiang2024}, the DSI framework can be used to predict spatio-temporal simulation responses relevant to GCS. Building on their approach, here $\left(\textbf{d}_{\text{full}}\right)_i$ is a vector containing ``flattened'' spatio-temporal fields of QoI in both the historical and prediction periods. We can separate $\left(\textbf{d}_{\text{full}}\right)_i$ into two components: $\left(\textbf{d}_{\text{hm}}\right)_i \in \mathbb{R}^{N_\text{hm} \times 1}$, which contains simulation results in the historical period, and $\left(\textbf{d}_{\text{pred}}\right)_i \in \mathbb{R}^{N_\text{pred} \times 1}$, denoting results for the prediction period. Thus, $\left(\textbf{d}_{\text{full}}\right)_i$ can be expressed as
\begin{equation}
    \left(\textbf{d}_{\text{full}}\right)_i = \left[(\mathbf{d}_\text{hm})_i^T,  (\mathbf{d}_\text{pred})_i^T\right]^T.
\end{equation}

Let $N_\text{t,hm}$ and $N_\text{t,pred}$ denote the number of time steps in the historical and prediction periods. The number of time steps considered in DSI differ from the number of time steps in the GEOS simulations, $N_\text{t,sim}$ (typically $N_\text{t,hm}+N_\text{t,pred} \ll N_\text{t,sim}$). In this study, the QoIs comprise four fields: pressure, shear stress, effective normal stress, and strain. Note that when we refer to strain, we always mean the $zz$-component of the strain tensor (which is the relevant component in this work). Although we focus only on $\sigma_n'$ and $\tau$ along the faults, these stress quantities are treated as 3D fields (similar to pressure and strain), with nonzero values only in the fault cells. This approach is used to facilitate VAE training. Because each field is computed for $N_\text{b}$ grid blocks, we have $N_\text{hm} = 4 N_\text{b} N_\text{t,hm}$, $N_\text{pred} = 4 N_\text{b} N_\text{t,pred}$, and $N_\text{full} = 4 N_\text{b} \left(N_\text{t,hm} +  N_\text{t,pred}\right)$. 

We generate a synthetic ``true'' model $\left(\textbf{m}_\text{true}\right)$ by randomly generating a permeability and porosity field and then sampling geomechanical and fault parameters from the ranges specified in Table~\ref{Table: realization params & prior params}. The true data $\left(\textbf{d}_\text{full}\right)_\text{true} \in \mathbb{R}^{N_\text{full} \times 1}$ are then generated via simulation, i.e., $\left(\textbf{d}_\text{full}\right)_\text{true} = g\left(\textbf{m}_\text{true}\right)$.
The true data at monitoring well locations during the historical period, represented by $\textbf{d}_\text{true} \in \mathbb{R}^{N_\text{obs} \times 1}$, can be expressed as
\begin{equation}
    \textbf{d}_\text{true} = H \left(\textbf{d}_\text{full}\right)_\text{true},
\label{Eq: dtrue = H dfulltrue}
\end{equation}
where $H \in \mathbb{R}^{N_\text{obs} \times N_\text{full}}$ is a selection matrix that extracts the historical monitoring data from the full true data vector. Observed data $\textbf{d}_\text{obs} \in \mathbb{R}^{N_\text{obs} \times 1}$ are generated by adding observation error (noise), denoted $\boldsymbol{\epsilon} \in \mathbb{R}^{N_\text{obs} \times 1}$, to $\textbf{d}_\text{true}$, 
\begin{equation}
    \textbf{d}_\text{obs} = \textbf{d}_\text{true} + \boldsymbol{\epsilon} = H \left(\textbf{d}_\text{full}\right)_\text{true} + \boldsymbol{\epsilon}.
\label{Eq: dobs = H dfulltrue + epsilon}
\end{equation}
Here we assume $\boldsymbol{\epsilon}$ follows a Gaussian distribution with independent components, i.e., $\boldsymbol{\epsilon} \sim \mathcal{N}\left(\mathbf{0}, C_\text{D}\right)$, where $C_\text{D}$ is a diagonal matrix with elements containing the standard deviation of the data errors. Model error can also be included in $C_\text{D}$ but this is not considered in this study.

Under the Bayesian framework, the posterior probability density function (PDF) of the data variables $\textbf{d}_\text{full}$ given the observation data $\textbf{d}_\text{obs}$ is expressed as
\begin{equation}
    p\left(\textbf{d}_\text{full} | \textbf{d}_\text{obs}\right) = \frac{p\left(\textbf{d}_\text{obs} | \textbf{d}_\text{full}\right) p\left(\textbf{d}_\text{full}\right)}{p\left(\textbf{d}_\text{obs}\right)} \propto p\left(\textbf{d}_\text{obs} | \textbf{d}_\text{full}\right) p\left(\textbf{d}_\text{full}\right).
\label{Eq: p(dfull_dobs)}
\end{equation}
Here $p\left(\textbf{d}_\text{obs} | \textbf{d}_\text{full}\right)$ is the likelihood function, $p\left(\textbf{d}_\text{full}\right)$ represents the prior PDF, and $p\left(\textbf{d}_\text{obs}\right)$ represents the evidence. The direct use of the $\textbf{d}_\text{full}$ vector in DSI can pose challenges, since $\textbf{d}_\text{full}$ is typically high-dimensional and its distribution is non-Gaussian. However, using the VAE introduced later, we obtain low-dimensional, approximately Gaussian-distributed latent vectors $\boldsymbol{\xi} \in \mathbb{R}^{N_\text{l} \times 1}$ that represent $\textbf{d}_\text{full}$ (here $N_\text{l}$ indicates the dimension of the VAE latent space). The relationship between the latent variables $\boldsymbol{\xi}$ and data variables $\textbf{d}_\text{full}$ can be expressed as
\begin{equation}
     {\textbf{d}}_\text{full} \approx \widehat{\textbf{d}}_\text{full} = f\left(\boldsymbol{\xi}\right),
\label{Eq: dfull&xi}
\end{equation}
where $\widehat{\textbf{d}}_\text{full} \in \mathbb{R}^{N_\text{full} \times 1}$ represents the data variables reconstructed from the latent variables. Eq.~\ref{Eq: p(dfull_dobs)} can then be written as 
\begin{equation} \label{eq:pxi}
    p(\boldsymbol{\xi} | \textbf{d}_{\text{obs}}) \propto p\left(\textbf{d}_\text{obs} | \boldsymbol{\xi}\right) p\left(\boldsymbol{\xi}\right).
\end{equation}

When the prior distribution of the latent variables $\boldsymbol{\xi}$ and observation error $\boldsymbol{\epsilon}$ are both Gaussian, Eq.~\ref{eq:pxi} can be expressed as
\begin{equation}
    p(\boldsymbol{\xi} | \mathbf{d}_{\text{obs}}) \propto \exp \left( -\frac{1}{2} \left( H f(\boldsymbol{\xi}) - \mathbf{d}_{\text{obs}} \right)^T C_\text{D}^{-1} \left( H f(\boldsymbol{\xi}) - \mathbf{d}_{\text{obs}} \right) - \frac{1}{2} \boldsymbol{\xi}^T \boldsymbol{\xi} \right).
\end{equation}
Our goal now is to generate posterior samples for $\boldsymbol{\xi}$, which can be decoded to provide $\widehat{\textbf{d}}_\text{full}$. We apply an ensemble smoother with multiple data assimilation (ESMDA) method to perform this posterior sampling. The ESMDA method updates the latent variables using the following equation \citep{lima2020data,jiang2021,jiang2024}:
\begin{equation}
    \boldsymbol{\xi}_j^{k+1} = \boldsymbol{\xi}_j^k + C_{{\xi}, \widehat{\mathrm{d}}_{\text{hm}}^{{\text{mon}}}}^{k} \left( C_{\widehat{\mathrm{d}}_{\text{hm}}^{{\text{mon}}}}^{k} + \alpha_k C_\text{D} \right)^{-1} \left( \mathbf{d}_{\text{obs}} + \sqrt{\alpha_k} \, \textbf{e}_j^k - \left( \widehat{\mathbf{d}}_{\text{hm}}^{{\text{mon}}} \right)_{j}^{k} \right),
\label{Eq: xi_jk+1=xi_jk+...}
\end{equation}
where $j = 1, ..., N_\mathrm{e}$ and $k = 1, ..., N_\mathrm{a}$, with $N_\mathrm{e}$ denoting the ensemble size and $N_\mathrm{a}$ the number of data assimilation steps. Here $\left(\widehat{\mathbf{d}}_{\text{hm}}^{{\text{mon}}}\right)_j^k \in \mathbb{R}^{N_\text{obs} \times 1}$ represents the predicted monitoring data (without noise) during the historical period, obtained by applying the selection matrix $H$ to the predicted data variables: 
\begin{equation}
    \left(\widehat{\mathbf{d}}_{\text{hm}}^{{\text{mon}}}\right)_j^k = H f\left(\boldsymbol{\xi}_j^k\right).
\label{Eq: dhmmon_jk=Hf(xi_jk)}
\end{equation}
The matrix $C_{{\xi}, \widehat{\mathrm{d}}_{\text{hm}}^{{\text{mon}}}} \in \mathbb{R}^{N_\text{l} \times N_\text{obs}}$ represents the cross-covariance matrix between $\boldsymbol{\xi}$ and $\widehat{\mathbf{d}}_{\text{hm}}^{{\text{mon}}}$, and $C_{\widehat{\mathrm{d}}_{\text{hm}}^{{\text{mon}}}} \in \mathbb{R}^{N_\text{obs} \times N_\text{obs}}$ denotes the covariance of $\widehat{\mathbf{d}}_{\text{hm}}^{{\text{mon}}}$. Here $\alpha_k$ is the inflation coefficient used at iteration $k$ (which should satisfy $\sum_{k=1}^{N_\mathrm{a}} \alpha_k^{-1} = 1$), and $\textbf{e}_j^k \sim \mathcal{N}\left(\mathbf{0}, C_\text{D}\right)$ is a random perturbation added to the observations.

In addition to providing posterior predictions of the QoIs, the DSI framework also allows us to predict key geomechanical and fault parameters. Here we consider $\log_{10}\left(k_\mathrm{f}^1 / k\right)$ and $\log_{10}\left(k_\mathrm{f}^2 / k\right)$, Young's modulus, $E$, Poisson ratio, $\nu$, Biot coefficient, $\alpha$, and the initial stress coefficient, $\gamma$. These parameters are incorporated into the latent variables $\boldsymbol{\xi}$ to allow for joint inversion. The variables updated during history matching, $\left(\boldsymbol{\xi}^{\text{jt}} \in \mathbb{R}^{\left(N_\text{l}+6\right) \times 1}\right)$, can be expressed as
\begin{equation}
    \boldsymbol{\xi}^{\text{jt}}=\left[\boldsymbol{\xi}^T, \log_{10}\left(k_\mathrm{f}^1 / k\right), \log_{10}\left(k_\mathrm{f}^2 / k\right), E, \nu, \alpha, \gamma\right]^T,
\end{equation}
where the superscript jt denotes joint inversion. Eq.~\ref{Eq: xi_jk+1=xi_jk+...} is now written as
\begin{equation}
    \left(\boldsymbol{\xi}^\text{jt}\right)_j^{k+1} = \left(\boldsymbol{\xi}^\text{jt}\right)_j^k + C_{{\xi}^\text{jt}, \widehat{\mathrm{d}}_{\text{hm}}^{{\text{mon}}}}^{k} \left( C_{\widehat{\mathrm{d}}_{\text{hm}}^{{\text{mon}}}}^{k} + \alpha_k C_\text{D} \right)^{-1} \left( \mathbf{d}_{\text{obs}} + \sqrt{\alpha_k} \, \textbf{e}_j^k - \left( \widehat{\mathbf{d}}_{\text{hm}}^{{\text{mon}}} \right)_{j}^{k} \right).
\end{equation}
Eq.~\ref{Eq: dhmmon_jk=Hf(xi_jk)} retains its original form, with $\boldsymbol{\xi}_j^k$ now corresponding to the first $N_\text{l}$ components of $\left(\boldsymbol{\xi}^\text{jt}\right)_j^{k}$.

In this study, the historical period includes time steps at 2, 4, 6, and 8~years, and the prediction period includes a single time of 50~years. Accordingly, we have $N_\text{t,hm}=4$ and $N_\text{t,pred}=1$. The monitoring data consist of pressure and strain ($zz$ component) observations within the storage aquifer during the historical period. For the Gaussian observation noise, the standard deviation of pressure is specified as 0.1~MPa \citep{jiang2024}, and the standard deviation of strain is defined as 10\% of the mean strain at the monitoring well locations over all recorded time steps~\citep{tolstukhin2014geologically}.

For history matching, we set $N_\mathrm{e} = 400$, $N_\mathrm{a} = 4$, $\alpha_1 = 9.33$, $\alpha_2 = 7.0$, $\alpha_3 = 7.0$, and $\alpha_4 = 2.0$ for all iterations. We also explored data assimilation using 10~iterations (and $\alpha_k = 10$), though this gave comparable results to the setup with 4~iterations. Once the posterior predictions of the latent variables are obtained, we generate posterior predictions for the data variables through application of Eq.~\ref{Eq: dfull&xi}.

\subsection{VAE for data parameterization}
\label{Subsec: VAE for data parameterization}

\textcolor{black}{The spatio-temporal data variables $\textbf{d}_\text{full}$ are high-dimensional and the distribution of any particular data variable (over the $N_\mathrm{e}$ ensemble members) is, in general, non-Gaussian. The VAE parameterization applied here, in which ${\textbf{d}}_\text{full} \in \mathbb{R}^{N_\text{full} \times 1}$ is represented in terms of a latent variable $\boldsymbol{\xi} \in \mathbb{R}^{N_\text{l} \times 1}$ that is approximately Gaussian, with $N_\text{l}\ll N_\text{full}$, addresses both these challenges. Specifically, the dimension reduction achieved here is very substantial. The dimension of ${\textbf{d}}_\text{full}$, not counting the padding of the stress variables from 2D to 3D, is 739,200, while the dimension of $\boldsymbol{\xi}$ is only 256. It is thus much more efficient to perform history matching with $\boldsymbol{\xi}$ than with ${\textbf{d}}_\text{full}$. In addition, many history matching methods, including ESMDA, involve assumptions of Gaussianity. If these assumptions are not satisfied, posterior (history matched) samples, which in our case correspond to data vectors, may exhibit nonphysical behaviors, such as spurious oscillations. The VAE constructed here provides (approximately) Gaussian-distributed $\boldsymbol{\xi}$, consistent with ESMDA assumptions, so posterior results do not display nonphysical behavior. We note finally that the use of the VAE parameterization in data-space history matching (DSI) is analogous to the use of geomodel parameterization in model-based history matching. \citet{difederico2025}, for example, developed a latent diffusion model to represent channelized (non-Gaussian) geological realizations, and then used the Gaussian latent variable in history matching in place of the physical geomodel parameters.}

\textcolor{black}{Models other than VAE could also be applied for this parameterization. \citet{jiang2024} used an adversarial autoencoder (AAE) model consisting of 3D ConvLSTM layers to parameterize pressure and saturation fields in their DSI implementation. AAEs have the advantage of more strongly enforcing a Gaussian distribution on the latent space. However, they rely on adversarial training, which can be challenging to stabilize, prone to mode collapse, and sensitive to hyperparameter tuning. We initially implemented an AAE model with an architecture similar to that of \citet{jiang2024}, but found that the training process was unstable and convergence was difficult to achieve. These problems were not encountered with the VAE developed in this work.} 

The VAE, like the AAE, is based on stacked 3D ConvLSTM layers. It includes an encoder-decoder architecture that regularizes the latent space to follow a Gaussian distribution. The VAE regularizes the latent space by directly minimizing the Kullback-Leibler (KL) divergence between the learned distribution and the desired Gaussian distribution during training~\citep{kingma2013auto}. \textcolor{black}{This approach achieves a balance between reconstruction accuracy and latent space regularization without the need for adversarial optimization, thus providing computational stability and efficiency.}

\begin{figure}
    \centering
    \includegraphics[width=\textwidth]{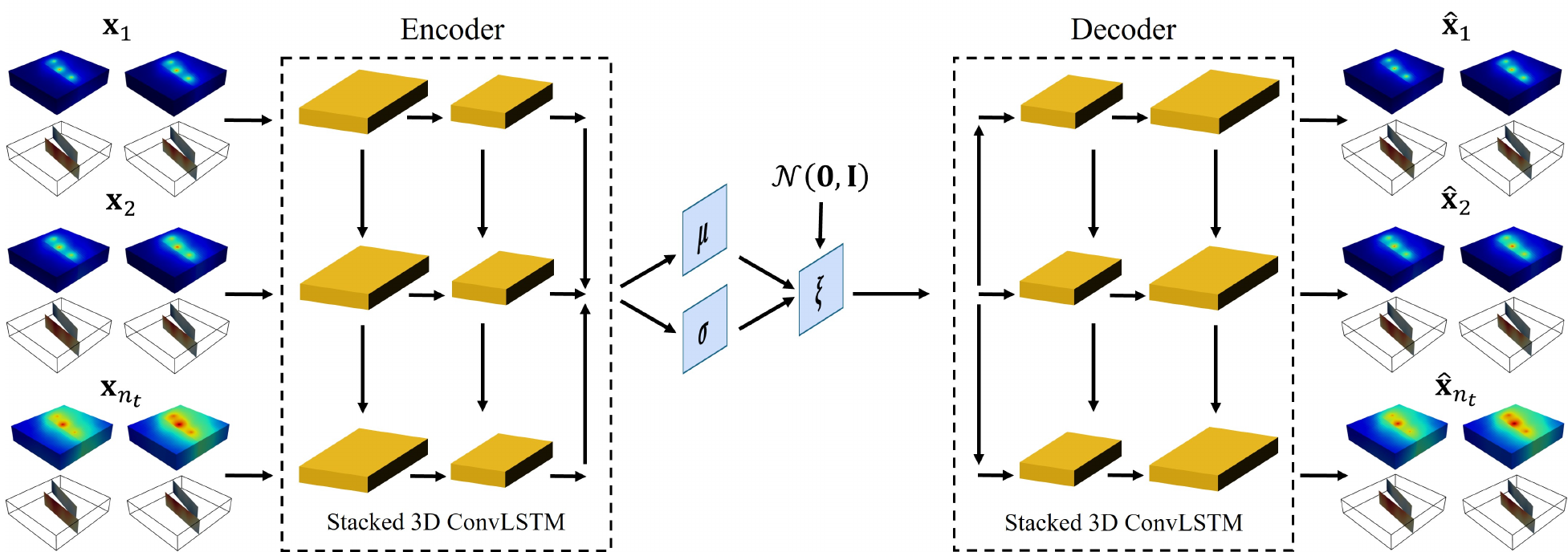}
    \caption{Schematic of the VAE architecture with stacked 3D ConvLSTM layers.}
    \label{Fig: vae}
\end{figure}

The VAE model is designed to learn (approximately) Gaussian-distributed low-dimensional representations for the high-dimensional pressure, strain, and stress fields at a set of time steps. These low-dimensional representations $\left(\boldsymbol{\xi}\right)$ are used in the history matching process, as described in Section~\ref{Subsec: DSI formulation}. The architecture of the VAE is illustrated in Fig.~\ref{Fig: vae}. It consists of an encoder-decoder structure, with both modules composed of multiple 3D ConvLSTM layers stacked in sequence. The pressure, strain, and stress states of the system at a given time step are denoted by $\boldsymbol{\textbf{x}}_i$, where $i=1, 2, \ldots, n_t$, with $n_t$ the total number of time steps considered in the VAE. The corresponding system states reconstructed by the VAE are indicated by $\hat{\textbf{x}}_i$. The parameterization process $f_\text{encoder}$ and the reparameterization process $f_\text{decoder}$ can be expressed as
\begin{equation}
    \boldsymbol{\xi}=f_\text{encoder}\left(\mathbf{d}_\text{full}; \boldsymbol{\theta}_\text{encoder}\right), \quad \widehat{\textbf{d}}_\text{full}=f_\text{decoder}\left(\boldsymbol{\xi}; \boldsymbol{\theta}_\text{decoder}\right),
\end{equation}
where $\boldsymbol{\theta}_\text{encoder}$ and $\boldsymbol{\theta}_\text{decoder}$ denote the trainable parameters associated with the encoder and decoder.

In the encoder, each ConvLSTM layer applies 3D convolutions that progressively reduce the spatial resolution. The final encoder layer condenses the spatial and temporal information, and then connects to a low-dimensional latent vector. This latent vector is parameterized in terms of (learned) means and variances to facilitate the probabilistic framework of the VAE \citep{kingma2013auto}. The decoder mirrors this structure, with stacked ConvLSTM layers gradually reconstructing the original data from the latent space. As we will see, the decoder is able to generate high-quality reconstructions that accurately capture both spatial and temporal characteristics of the data. Details of the ConvLSTM and VAE architectures are provided in SI.

Prior to VAE training, we apply z-score normalization to the pressure and strain fields and to the nonzero entries in the stress fields. The detailed expressions for these normalizations are given in SI. The VAE is then trained by minimizing a loss function $\mathcal{L}\left(\boldsymbol{\theta}_\text{encoder}, \boldsymbol{\theta}_\text{decoder}; \textbf{d}_\text{full} \right)$ that combines reconstruction loss $\mathcal{L}_\text{recon}$ and the KL divergence loss $\mathcal{L}_\text{KL}$ \citep{kingma2013auto}. The reconstruction loss quantifies the mismatch between the reconstructed and input fields, while the KL divergence quantifies the mismatch between the distribution of $\boldsymbol{\xi}$ and a standard normal distribution. The total loss function $\mathcal{L}$ is given by
\begin{equation} \label{eq:loss}
    \mathcal{L}\left(\boldsymbol{\theta}_\text{encoder}, \boldsymbol{\theta}_\text{decoder}; \textbf{d}_\text{full} \right) =  \omega\mathcal{L}_\text{recon} + \mathcal{L}_\text{KL},
\end{equation}
where $\omega$ is a parameter that balances reconstruction quality and latent space regularization. The quantities $\mathcal{L}_\text{recon}$ and $\mathcal{L}_\text{KL}$ are computed as
\begin{subequations}
\begin{align}
    \mathcal{L}_\text{recon} &= \frac{1}{N_\text{train}} \sum_{j=1}^{N_\text{train}} \left\| \left(\textbf{d}_\text{full}^\text{norm}\right)^j - \left( \widehat{\mathbf{d}}_{\text{full}}^{{\text{norm}}} \right)^j \right\|_2 ^2, \\
    \mathcal{L}_\text{KL} &= \frac{1}{N_\text{train}} \sum_{j=1}^{N_\text{train}} \left[ -\frac{1}{2} \sum_{i=1}^{N_\text{l}} \left(1 + 2\log\left(\sigma_i^j\right) - \left(\mu_i^j\right)^2 - \left(\sigma_i^j\right)^2\right) \right].
\end{align}
\end{subequations}
Here $N_\text{train}$ is the number of training samples, $\left(\textbf{d}_\text{full}^\text{norm}\right)^j$ represents the normalized $\left(\textbf{d}_\text{full}\right)^j$ for training sample $j$, $\left( \widehat{\mathbf{d}}_{\text{full}}^{{\text{norm}}} \right)^j$ is the corresponding VAE output, and $\mu_i^j$ and $\sigma_i^j$ denote the mean and standard deviation of component $i$ of the latent variable $\boldsymbol{\xi}$ for training sample $j$.

\textcolor{black}{Up to this point, we have described the DSI procedure and the VAE architecture and loss formulation. In the next section we will evaluate the reconstruction performance of the VAE, with the goal of demonstrating that the low-dimensional latent representations $\boldsymbol{\xi}$ can successfully capture the key spatial and temporal features of the input fields. The generative quality of the VAE will also be examined to evaluate its ability to produce realistic new fields from Gaussian latent vector inputs.}

\section{\textcolor{black}{Verification of VAE reconstruction and generation capabilities}} \label{Sec: VAE reconstruction results}

\textcolor{black}{The VAE model described in Section~\ref{Subsec: VAE for data parameterization} will now be evaluated in terms of reconstruction accuracy and generative quality. This enables us to assess whether the learned latent representations effectively capture the spatio-temporal fields of interest, and whether they are suitable for use within the DSI framework}. To train the VAE model, 1500 realizations are first generated (as described in Section~\ref{Subsec: Faulted geomodel}) and then simulated. Of these, 1200 realizations are used as the training set, 150 as the validation set, and 150 as the test set. The input and output of the VAE comprise the pressure, strain ($zz$ component), and effective normal stress and shear stress fields at seven time steps (2, 4, 6, 8, 20, 36, and 50~years). These time steps cover both the history matching period (2, 4, 6, and 8~years) and the prediction period (50~years). The intermediate steps (20 and 36~years) are included to improve accuracy in temporal evolution. 

The VAE is trained using a batch size of 8 for 600~epochs, starting with an initial learning rate of $7.5\times10^{-4}$. We employ a stage-wise KL annealing strategy by progressively decreasing $\omega$ (in Eq.~\ref{eq:loss}) during training. A value of $\omega = 10^5$ is used for the first 100~epochs, and the final value of $\omega$ (used for the last 200~epochs) is 100. The VAE training requires around 12~hours on a Nvidia A100 GPU.

\begin{figure}
    \centering
    \includegraphics[width=0.5\textwidth]{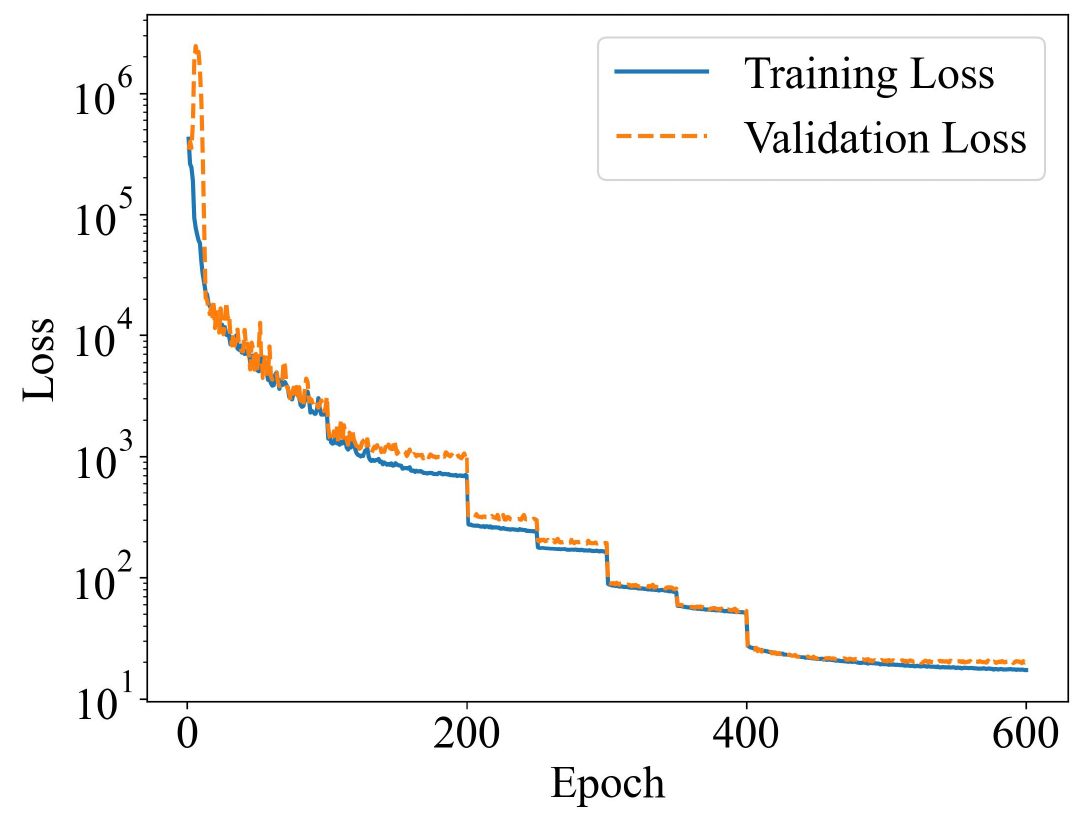}
    \caption{\textcolor{black}{Training (solid blue curve) and validation (dashed orange curve) losses of the VAE model.}}
\label{Fig: loss curves}
\end{figure}

\textcolor{black}{The training and validation losses over all 600~epochs are presented in Fig.~\ref{Fig: loss curves}. Both curves clearly decrease and stay very close to each other, indicating stable training without obvious overfitting. The step-like decline in the loss is due to the effect of the KL annealing schedule. The results in Fig.~\ref{Fig: loss curves} demonstrate that the VAE model is stable and convergent.}

\begin{figure}
    \centering
    \includegraphics[width=0.8\textwidth]{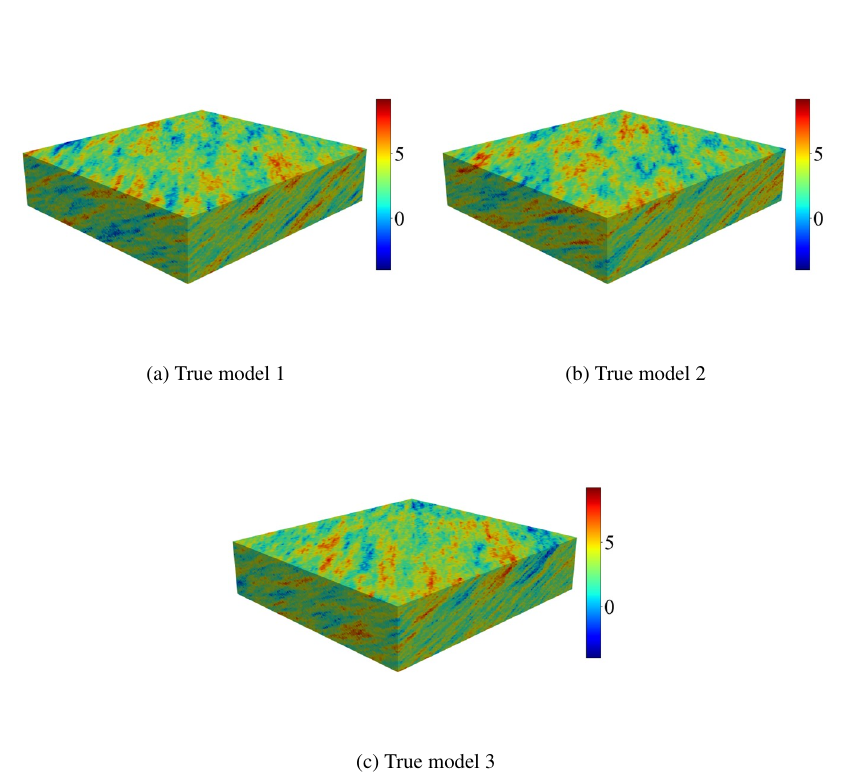}
    \caption{Log-permeability ($\log_e k_x$, with $k_x$ in md) realizations for the three true models used in history matching.}
\label{Fig: 3 true model realizations}
\end{figure}

\begin{figure}
    \centering
    \includegraphics[width=1\textwidth]{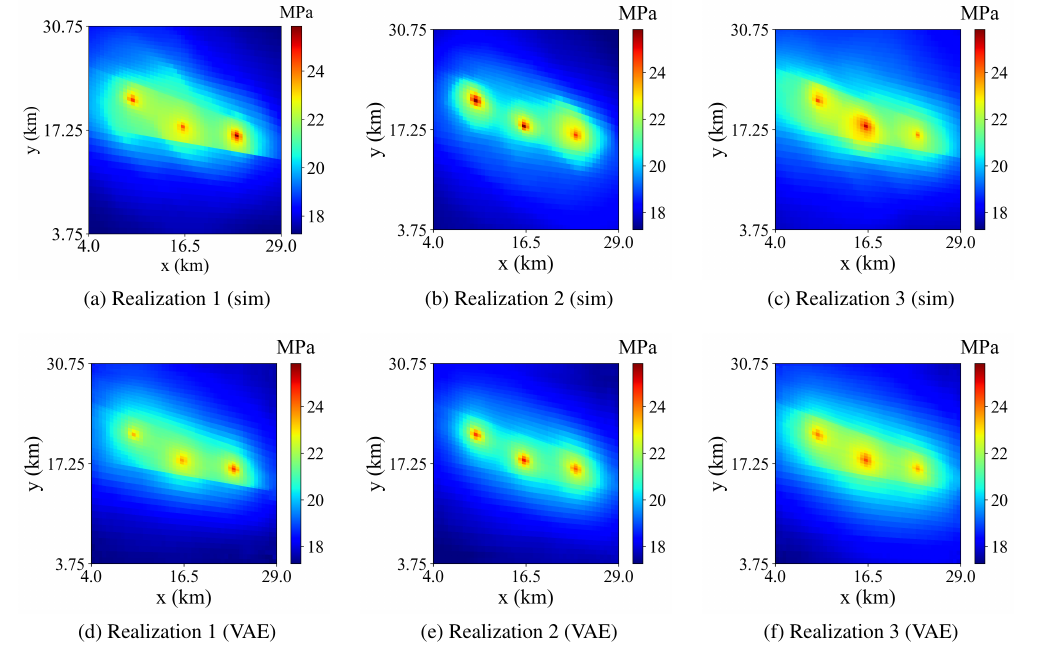}
    \caption{Pressure fields from the GEOS simulations (upper row) and VAE reconstructions (lower row) in the top layer of the storage aquifer at year~50.}
\label{Fig: VAE recon pressure}
\end{figure}

\begin{figure}
    \centering
    \includegraphics[width=1\textwidth]{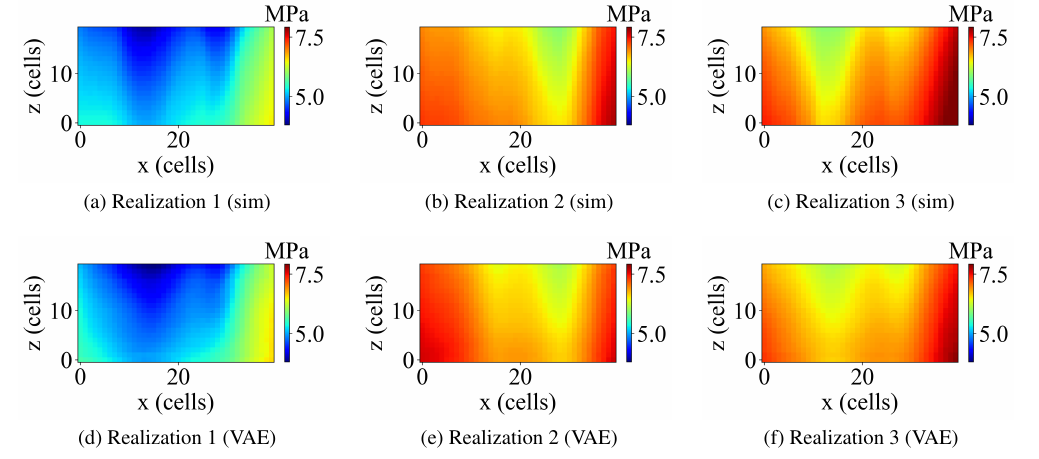}
    \caption{Effective normal stress along Fault~1 from the GEOS simulations (upper row) and VAE reconstructions (lower row) in the storage aquifer at year~50.}
\label{Fig: VAE recon normal stress}
\end{figure}

The reconstruction performance of the trained VAE model is now assessed. Three realizations from the test set are selected, as shown in Fig.~\ref{Fig: 3 true model realizations}. These realizations correspond to the true models used for history matching in Section~\ref{Sec:DSI_results} and SI. A comparison between the reconstructed pressure fields and the reference simulation results for these three realizations is presented in Fig.~\ref{Fig: VAE recon pressure}. The reference pressure results are shown in the upper row and the reconstructions appear in the lower row. These results correspond to the top layer of the storage aquifer at year~50. Overall, the reconstructed pressure fields align closely with the reference simulation results, though some slight differences are apparent. Fig.~\ref{Fig: VAE recon normal stress} shows the effective normal stress along Fault~1 in the storage aquifer. These figures demonstrate strong agreement between the reconstructions and the reference simulation results. Significant differences in the stress fields between the three realizations are evident, indicating that the VAE can capture a range of responses. Additional reconstruction results for strain and shear stress are provided in the SI.

\textcolor{black}{We now quantify the relative errors of the reconstructed pressure, strain, effective normal stress, and shear stress fields. These errors, for test case $i$, are denoted as $\delta^{i}_{p}$, $\delta^{i}_{\varepsilon}$, $\delta^{i}_{\sigma_n^\prime}$ and $\delta^{i}_{\tau}$, respectively, and are computed as
\begin{samepage}
\begin{subequations}
\begin{alignat}{2}
    \delta^{i}_{p} &= \frac{1}{N_\text{b} n_t} \sum_{j=1}^{N_\text{b}} \sum_{t=1}^{n_t} \frac{\left| \hat{p}^{t}_{i,j} - p^{t}_{i,j} \right|} {p_{i,\max}^t - p_{i,\min}^t}, 
    &\qquad \delta^{i}_{\varepsilon} &= \frac{1}{N_\text{b} n_t} \sum_{j=1}^{N_\text{b}} \sum_{t=1}^{n_t} \frac{\left| \hat{\varepsilon}^{t}_{i,j} - \varepsilon^{t}_{i,j} \right|} {\varepsilon_{i,\max}^t - \varepsilon_{i,\min}^t}, \\
    \delta^{i}_{\sigma_n^\prime} &= \frac{1}{N_{\Omega_\text{f}} n_t} \sum_{j=1}^{N_{\Omega_\text{f}}} \sum_{t=1}^{n_t} \frac{\left| \left(\hat{\sigma_n^\prime}\right)^{t}_{i,j} - \left({\sigma_n^\prime}\right)^{t}_{i,j} \right|} {\left({\sigma_n^\prime}\right)_{i,\max}^t - \left({\sigma_n^\prime}\right)_{i,\min}^t}, 
    &\qquad \delta^{i}_{\tau} &= \frac{1}{N_{\Omega_\text{f}} n_t} \sum_{j=1}^{N_{\Omega_\text{f}}} \sum_{t=1}^{n_t} \frac{\left| \hat{\tau}^{t}_{i,j} - \tau^{t}_{i,j} \right|} {\tau_{i,\max}^t - \tau_{i,\min}^t}.
    \label{Eq: vae relative error for stress}
\end{alignat}
\end{subequations}
\end{samepage}
Here $\hat{p}^{t}_{i,j}$, $\hat{\varepsilon}^{t}_{i,j}$, $\left(\hat{\sigma_n^\prime}\right)^{t}_{i,j}$, and $\hat{\tau}^{t}_{i,j}$ represent the reconstructed pressure, strain, effective normal stress, and shear stress, respectively, for test case $i$, grid block $j$, and time step $t$. The corresponding reference values from the GEOS simulations are denoted as ${p}^{t}_{i,j}$, ${\varepsilon}^{t}_{i,j}$, $\left({\sigma_n^\prime}\right)^{t}_{i,j}$, and ${\tau}^{t}_{i,j}$. As noted earlier, $N_\text{b}$ is the number of grid blocks in the model and $n_t$ is the number of time steps considered in the VAE. For stress error computations, $N_{\Omega_\text{f}}$ refers to the number of grid blocks associated with the fault regions. }

\begin{figure}
    \centering
    \includegraphics[width=0.6\textwidth]{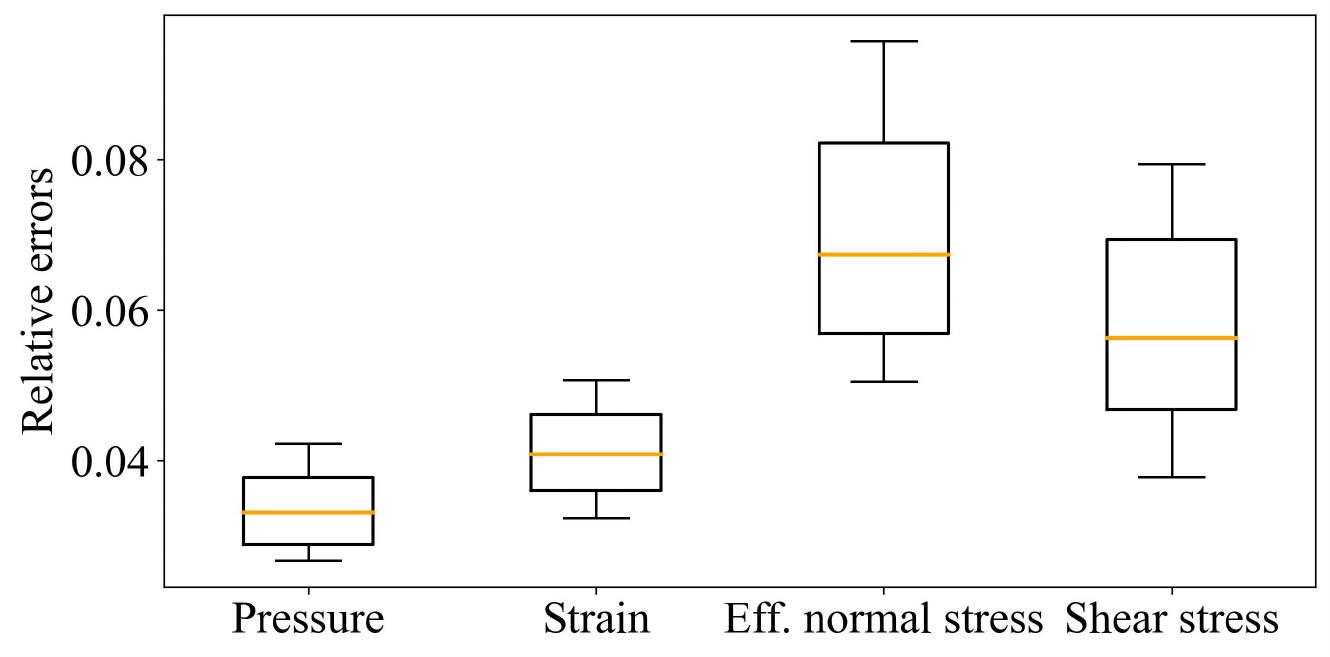}
    \caption{\textcolor{black}{Relative errors for VAE reconstruction results for the test set for pressure, strain, effective normal stress, and shear stress. Boxes display $\text{P}_{90}$, $\text{P}_{75}$, $\text{P}_{50}$, $\text{P}_{25}$ and $\text{P}_{10}$ percentile errors.}}
\label{Fig: VAE errors}
\end{figure}

\textcolor{black}{The relative errors over the 150 test cases are shown as box plots in Fig.~\ref{Fig: VAE errors}. In each box, the solid orange line indicates the median ($\text{P}_{50}$) error, the box boundaries show the $\text{P}_{75}$ and $\text{P}_{25}$ percentile errors, and the whiskers represent the $\text{P}_{90}$ and $\text{P}_{10}$ percentile errors. The median errors are approximately 0.033 for pressure and 0.041 for strain, while the errors for effective normal stress and shear stress are slightly higher, around 0.067 and 0.056, respectively. It is important to note that the normalizations applied in these computations lead to conservative (higher) error estimates. If the errors for effective normal stress and shear stress are instead normalized by their simulation values, i.e., using $\left({\sigma_n^\prime}\right)^{t}_{i,j}$ and ${\tau}^{t}_{i,j}$ in the denominators of Eq.~\ref{Eq: vae relative error for stress}, the resulting relative errors are about 0.019 for effective normal stress and 0.015 for shear stress.}

To evaluate the generative quality of the VAE, we next compare the statistical characteristics of the pressure and strain fields produced by the VAE with those obtained from numerical simulations. The results here correspond to data at monitoring well locations. The monitoring wells, denoted O1--O4, are shown in Fig.~\ref{Fig: realization, injectors, monitors, & fault mesh}(b). These locations are determined by minimizing the expected posterior uncertainty of a combination of QoIs (as described in SI). Each monitoring well collects pressure and strain data at each layer of the storage aquifer at all DSI time steps.

\begin{figure}
    \centering
    \includegraphics[width=1\textwidth]{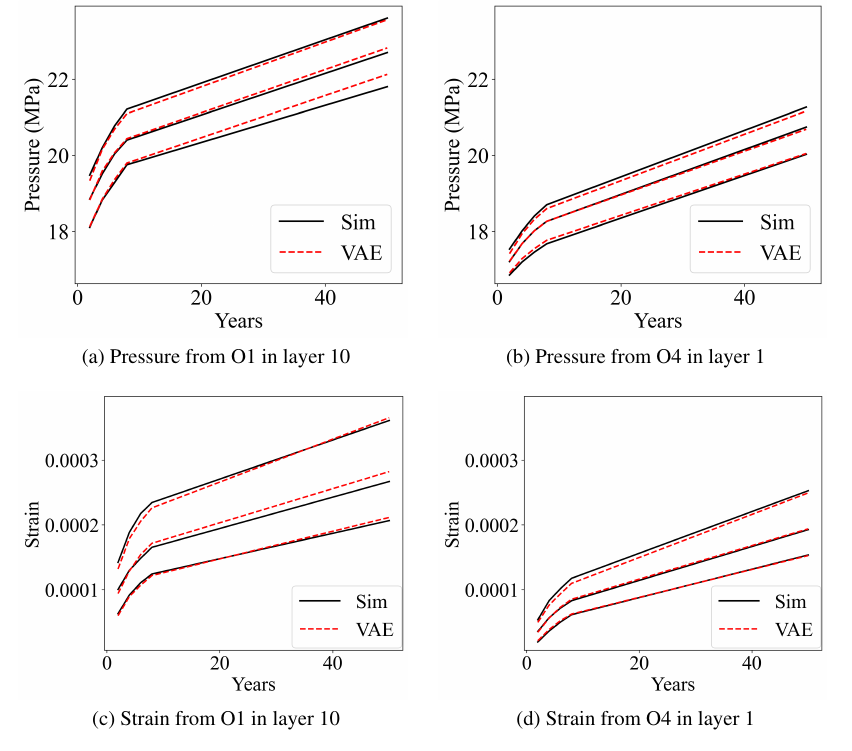}
    \caption{Pressure and strain statistics from GEOS simulations for test cases (black solid curves) and new realizations generated by the VAE (red dashed curves). The lower, middle, and upper curves represent the $\text{P}_{10}$, $\text{P}_{50}$, and $\text{P}_{90}$ responses.}
\label{Fig: VAE statistics}
\end{figure}

VAE-generated results are shown in Fig.~\ref{Fig: VAE statistics}. The new VAE data realizations are generated by randomly sampling 150 latent variables $\boldsymbol{\xi}$ from a 256-dimensional normal distribution, and then decoding them to produce the corresponding fields. For comparison, the numerical simulation results are taken from the 150 realizations in the test set. Results correspond to data from monitoring well O1 in layer~10 (middle of the storage aquifer) and O4 in layer~1 (top of the aquifer). Each figure shows the $\text{P}_{10}$, $\text{P}_{50}$, and $\text{P}_{90}$ (percentile) responses. The black solid lines represent the simulation results, while the red dashed lines show the VAE-generated outputs. The results demonstrate that the VAE effectively replicates the primary statistical characteristics of the simulation results, though slight differences are evident at some locations and time steps. In an overall sense, these results suggest the trained VAE can provide an efficient alternative to full-order GEOS simulations for our history matching problem.

\section{DSI-VAE history matching results}
\label{Sec:DSI_results}

We now present the history matching results obtained using the DSI-VAE framework. Three cases, each involving different true models, are considered (the true models are shown in Fig.~\ref{Fig: 3 true model realizations}). The geomechanical and fault parameters for these models are sampled from the prior distributions listed in Table~\ref{Table: realization params & prior params}. The parameter values for the three true models are provided in Table~\ref{Table: true model values}. With the trained VAE, a full DSI run (using ESMDA with $N_\mathrm{e} = 400$ and $N_\mathrm{a} = 4$) requires only about 7~minutes on a Nvidia A100 GPU. Recall that a single GEOS run takes $\sim$2~hours (using the AMD EPYC-7543 32-Core CPU).

\begin{table}
    \centering
    \caption{Geomechanical and fault parameters for three synthetic true models.}
    \begin{tabular}{ccccccc}
        \toprule
         & {$\log_{10}\left(k_\mathrm{f}^1 / k\right)$} & {$\log_{10}\left(k_\mathrm{f}^2 / k\right)$} & {$E$} & {$\nu$} & {$\alpha$} & {$\gamma$} \\
        \midrule
        True model~1 & -2.818 & -1.968 & 12.59~GPa & 0.262 & 0.960 & 0.981 \\
        True model~2 & -1.457 & -1.458 & 14.31~GPa & 0.271 & 0.859 & 0.236 \\
        True model~3 & -2.388 & -2.019 & 14.40~GPa & 0.290 & 0.838 & 0.259 \\
        \bottomrule
    \end{tabular}
\label{Table: true model values}
\end{table}

\begin{figure}
    \centering
    \includegraphics[width=1\textwidth]{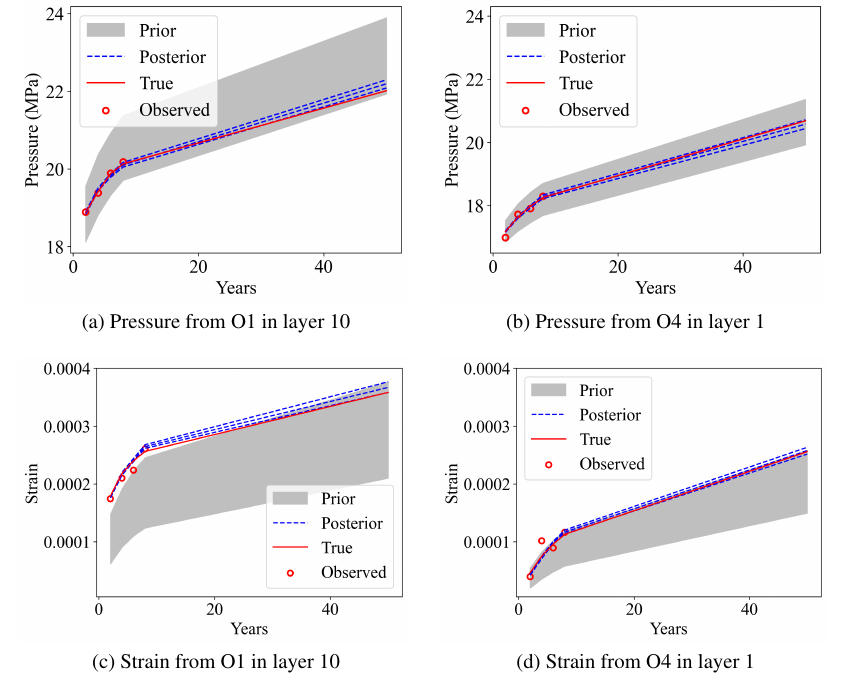}
    \caption{Pressure and strain ($zz$ component) results at monitoring locations for Case~1. Gray areas denote the prior $\text{P}_{10}$--$\text{P}_{90}$ range, red lines represent the true model response, red circles indicate observed data (which include noise), and dashed blue curves denote the posterior $\text{P}_{10}$, $\text{P}_{50}$, and $\text{P}_{90}$ predictions.}
\label{Fig: dsi pressure & strain obs}
\end{figure}

History matching results for pressure at monitoring wells O1 (layer~10) and O4 (layer~1), for Case~1, are shown in Fig.~\ref{Fig: dsi pressure & strain obs}(a) and (b). In Case~1, the data used for history matching are provided by simulating True model~1. The solid red lines represent the true pressure response (without noise) obtained from simulation, and the red circles correspond to the observed pressure values (with noise). The gray regions indicate the prior $\text{P}_{10}$--$\text{P}_{90}$ range and the dashed blue lines show the posterior $\text{P}_{10}$, $\text{P}_{50}$, and $\text{P}_{90}$ predictions. The posterior results demonstrate a very substantial reduction in uncertainty compared to the prior range. Moreover, at the monitoring well locations, the posterior predictions align well with the true pressure.

DSI results for strain at the same locations are shown in Fig.~\ref{Fig: dsi pressure & strain obs}(c) and (d). We again observe large reductions in uncertainty compared to the prior $\text{P}_{10}$--$\text{P}_{90}$ ranges, and accuracy relative to the true response. The increased (relative) noise level in strain compared to pressure is also evident. It is noteworthy that, although the observed data fall near the edge of the prior in some cases (outside $\text{P}_{90}$ in Fig.~\ref{Fig: dsi pressure & strain obs}(c) and near $\text{P}_{90}$ in Fig.~\ref{Fig: dsi pressure & strain obs}(d)), the DSI-VAE procedure is still able to provide effective data assimilation.
\textcolor{black}{In particular, at year~8 in Fig.~\ref{Fig: dsi pressure & strain obs}(c), the true strain corresponds to about the $\text{P}_{92}$ of the prior and the $\text{P}_{1}$ of the posterior. Although the posterior range for this quantity is very narrow, there is close agreement between the predictions and the true strain.}

\begin{figure}
    \centering
    \includegraphics[width=1\textwidth]{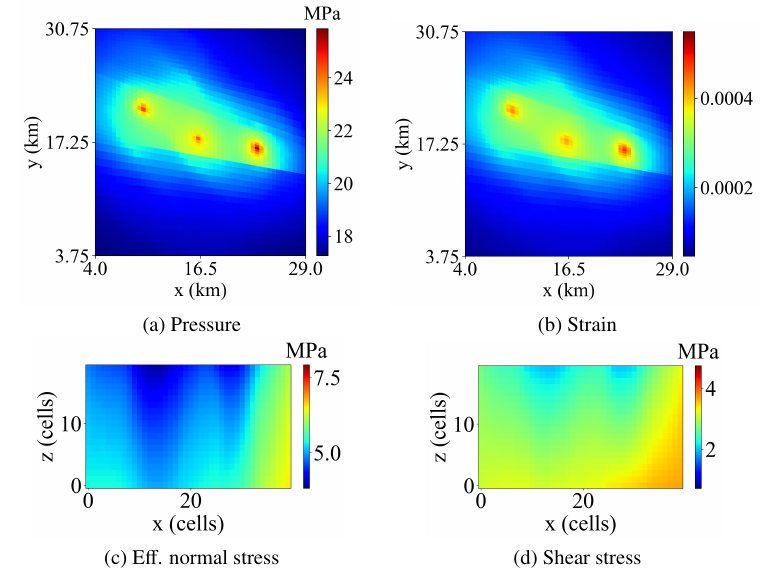}
    \caption{True pressure and strain ($zz$ component) fields in the top layer of the storage aquifer (a, b), and effective normal and shear stress along Fault~1 (c, d) in the storage aquifer, all at year~50 for Case~1.}
\label{Fig: all true fields}
\end{figure}

\begin{figure}
    \centering
    \includegraphics[width=1\textwidth]{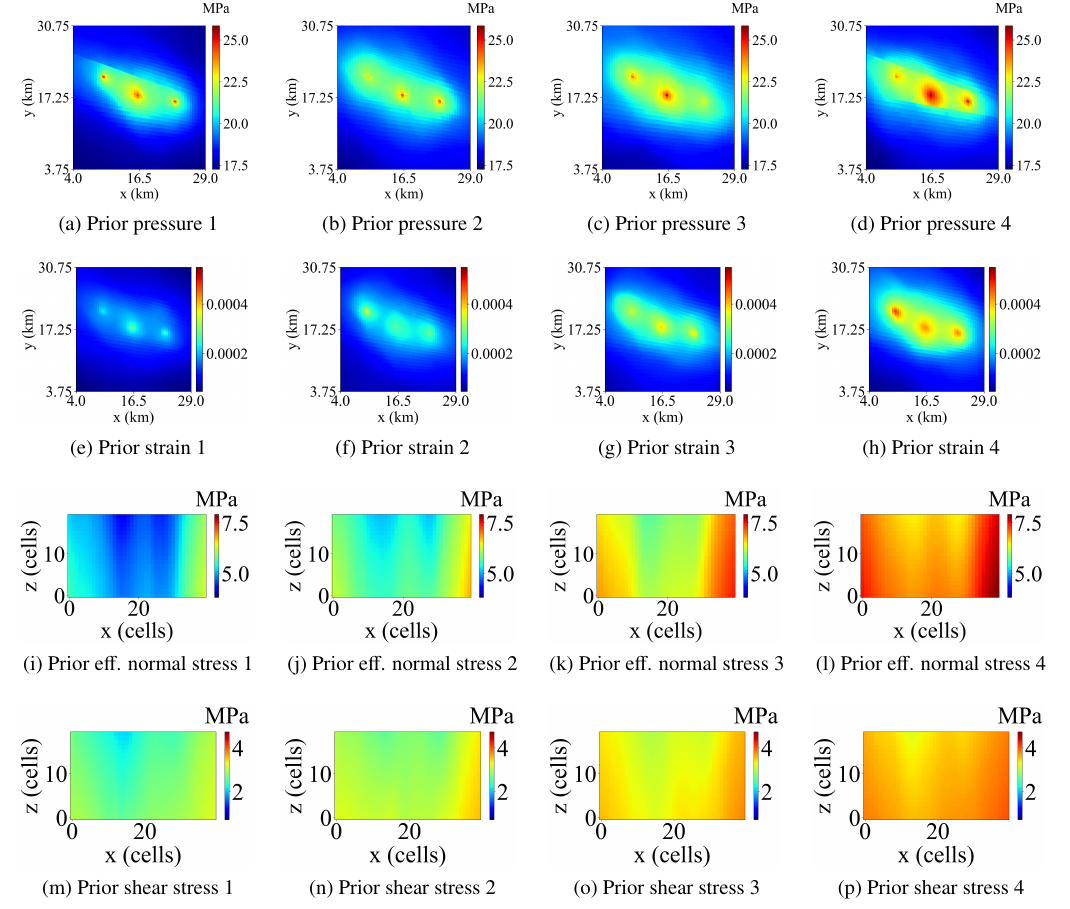}
    \caption{Representative prior pressure (first row) and $zz$ component of strain (second row) in the top layer of the storage aquifer at year~50. Representative prior effective normal stress (third row) and shear stress (fourth row) along Fault~1 in the storage aquifer at year~50.}
\label{Fig: representative priors}
\end{figure}

Next, we present representative prior and posterior pressure fields for Case~1. These representative fields are identified using the k-means and k-medoids method described in \citet{jiang2024}. With this approach, the k-means algorithm is used to group the prior or posterior pressure fields into four clusters, and the medoid of each cluster is selected using the k-medoids method. Each medoid is then taken to be a representative realization. The true fields for Case~1 appear in Fig.~\ref{Fig: all true fields}.

The first row of Fig.~\ref{Fig: representative priors} presents four such prior pressure fields, while the second row shows prior strain fields ($zz$ component). These maps are all for the top layer of the storage aquifer at year~50. The third and fourth rows of Fig.~\ref{Fig: representative priors} show prior effective normal stress and shear stress, respectively. The stress results are along Fault~1 within the storage aquifer at year~50. The prior results for all four QoI display large variations. The strain fields, for example, indicate different amounts of displacement over regions of varying size. The normal stress and shear stress fields also show a range of magnitudes and different spatial characteristics between realizations. These results thus illustrate the large range of uncertainty associated with the prior data variables.

\begin{figure}
    \centering
    \includegraphics[width=1\textwidth]{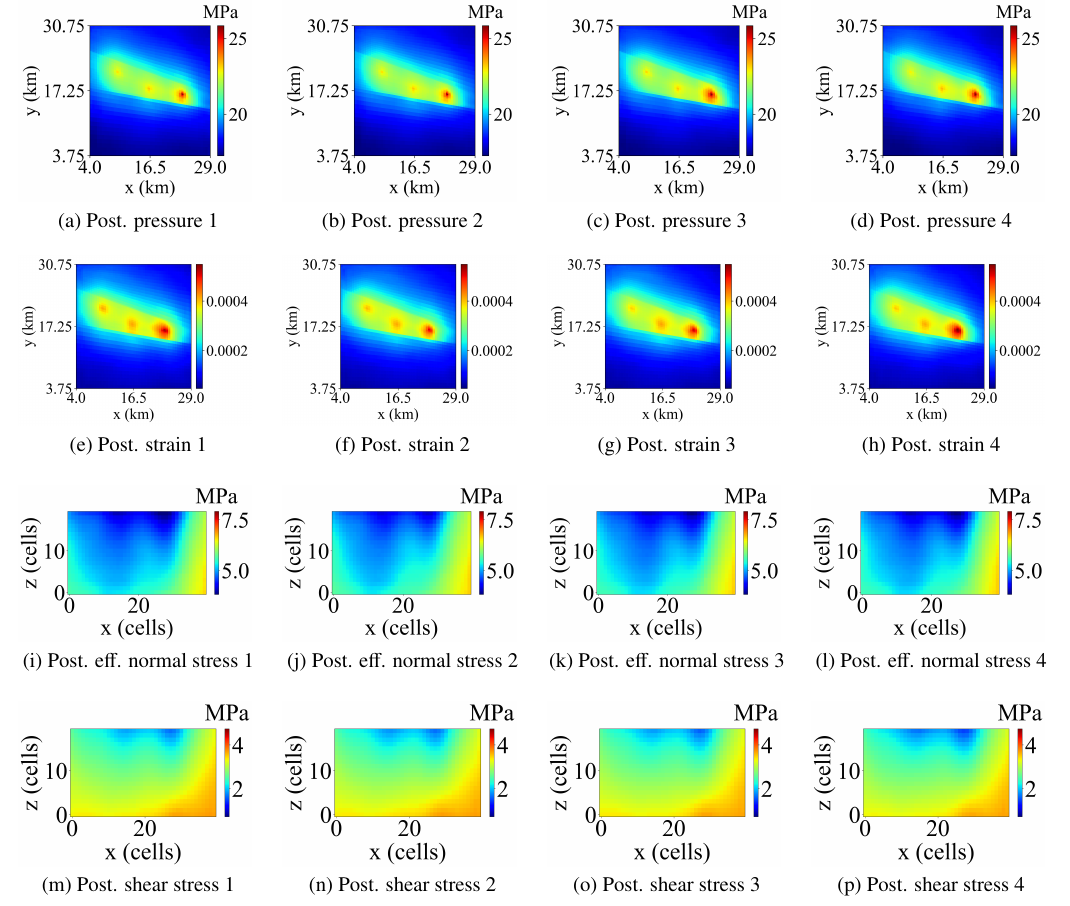}
    \caption{Representative posterior pressure (first row) and $zz$ component of strain (second row) in the top layer of the storage aquifer at year~50 for Case~1. Representative posterior effective normal stress (third row) and shear stress (fourth row) along Fault~1 in the storage aquifer at year~50 for Case~1. The corresponding true pressure, strain, effective normal stress, and shear stress results (based on True model~1) are shown in Fig.~\ref{Fig: all true fields}.}
\label{Fig: representative posteriors}
\end{figure}

Posterior results for the same four fields, again selected using the k-means and k-medoids method, are shown in Fig.~\ref{Fig: representative posteriors}. 
Immediately evident in each row of Fig.~\ref{Fig: representative posteriors} is that the representative posterior fields show much less variability than the prior fields. The shape of the region with notable strain (second row of Fig.~\ref{Fig: representative posteriors}), for example, is nearly the same in all cases, indicating significant uncertainty reduction relative to the prior results. The features apparent in the pressure and stress fields are also consistent from posterior realization to posterior realization. Comparing the fields in Fig.~\ref{Fig: representative posteriors} to those in Fig.~\ref{Fig: all true fields}, it is evident that the history matched DSI-VAE results closely resemble the true results for these important QoI.

\begin{figure}
    \centering
    \includegraphics[width=1\textwidth]{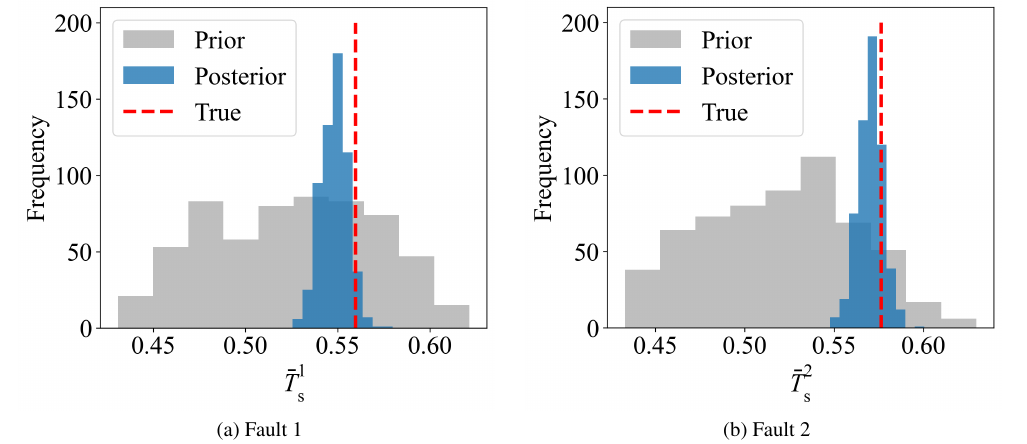}
    \caption{Posterior results (blue histograms) for the average fault slip tendencies $\bar{T}_\text{s}^1$ and $\bar{T}_\text{s}^2$ for Case~1 at year~50. Gray histograms show the prior distributions and the true values are indicated by the vertical red lines.}
\label{Fig: posterior slip tendencies for true model 1}
\end{figure}

\begin{figure}
    \centering
    \includegraphics[width=1\textwidth]{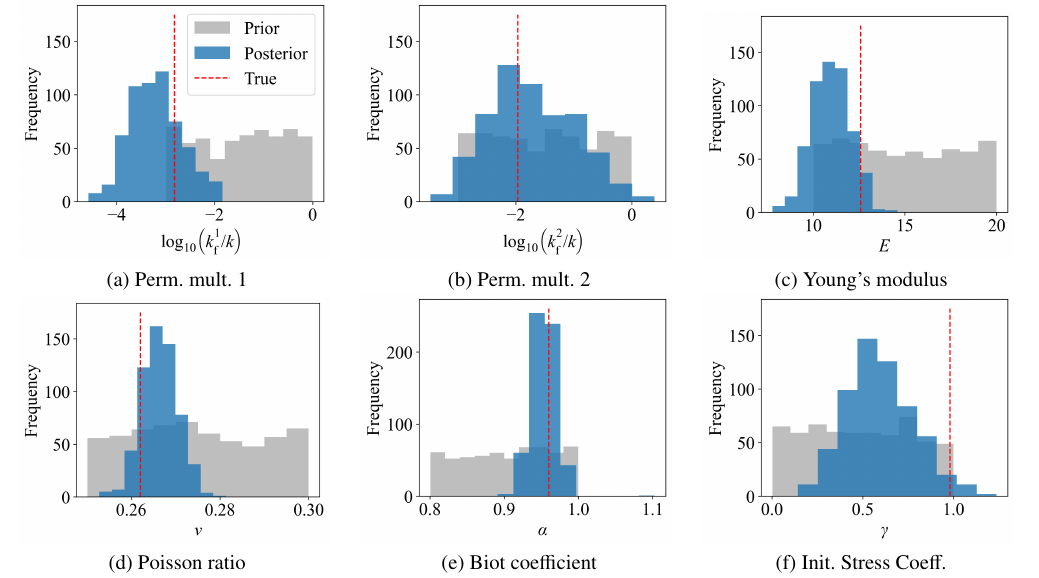}
    \caption{Posterior results (blue histograms) for $\log_{10}$ fault permeability multipliers, Young's modulus, Poisson ratio, Biot coefficient, and initial stress coefficient for Case~1. Gray histograms show the prior distributions and the true values are indicated by the vertical red lines.}
\label{Fig: posterior parameters for true model 1}
\end{figure}

Using the posterior predictions for shear stress and effective normal stress at year~50, we  calculate the posterior slip tendency for each cell along the fault using Eq.~\ref{Eq: Ts = tau/sigma_n'}. Results for average FST for the two faults at year~50, denoted $\bar{T}_\text{s}^1$ and $\bar{T}_\text{s}^2$, are presented in Fig.~\ref{Fig: posterior slip tendencies for true model 1}. These averages are computed over all cells along the fault. The gray and blue histograms represent the prior and posterior distributions, respectively, and the true values are shown as the vertical red lines. For this case, both $\bar{T}_\text{s}^1$ and $\bar{T}_\text{s}^2$ are relatively close to the threshold of 0.6 (discussed in Section~\ref{Subsec: Faulted geomodel}). The prior distributions exhibit fairly wide ranges, with $\bar{T}_\text{s}$ ranging from 0.43 to 0.63. In contrast, the posterior distributions are more narrowly concentrated around the true values, with most posterior predictions falling within the 0.53--0.58 range. This suggests that this case could experience fault slip if additional CO$_2$ is injected. Posterior results for $\bar{T}_\text{s}$ along the lines of those in Fig.~\ref{Fig: posterior slip tendencies for true model 1} might motivate an operator to modify the time frame or injection rates of the storage operation.

As discussed in Section~\ref{Subsec: DSI formulation}, in addition to posterior pressure, strain, and stress fields, DSI can also provide posterior estimates of key model parameters. Results for fault permeability multipliers, Young's modulus, Poisson ratio, Biot coefficient, and the initial stress coefficient for Case~1 are shown in Fig.~\ref{Fig: posterior parameters for true model 1}. Prior and posterior histograms are again shown in gray and blue, and the vertical red lines indicate the true values. Varying degrees of uncertainty reduction in these parameters are observed. Substantial uncertainty reduction is achieved for Young's modulus, Poisson ratio, and Biot coefficient (Fig.~\ref{Fig: posterior parameters for true model 1}(c), (d), and (e)), though wider posterior distributions are evident for the other parameters.

\begin{figure}
    \centering
    \includegraphics[width=1\textwidth]{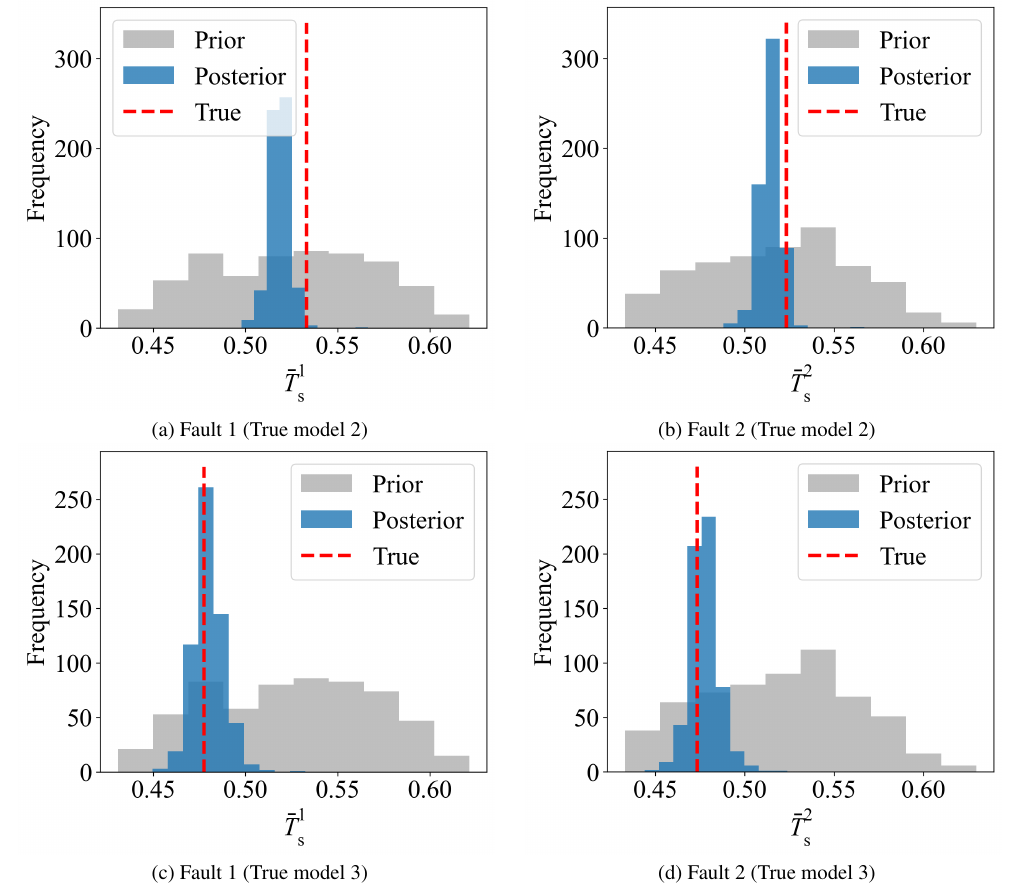}
    \caption{\textcolor{black}{Posterior results (blue histograms) for the average fault slip tendencies $\bar{T}_\text{s}^1$ and $\bar{T}_\text{s}^2$ for (a, b) Case~2 and (c, d) Case~3 at year~50. Gray histograms show the prior distributions and the true values are indicated by the vertical red lines.}}
\label{Fig: posterior slip tendencies for true models 2 and 3}
\end{figure}

\textcolor{black}{We now present a concise set of history matching results for two additional cases. The true models for Cases~2 and 3 are shown in Fig.~\ref{Fig: 3 true model realizations}(b) and (c), and the detailed model parameters are provided in Table~\ref{Table: true model values}. Posterior results for fault slip tendencies for these two cases are shown in Fig.~\ref{Fig: posterior slip tendencies for true models 2 and 3}. 
Young's modulus is larger in Cases~2 and~3 than in Case~1, while the Biot coefficient is smaller. A larger Young's modulus leads to a stiffer rock matrix and a lower Biot coefficient reduces the contribution of pore pressure to the total stress. In addition, fault permeabilities are generally (slightly) larger in Cases~2 and~3, which reduces pressure buildup and thus decreases fault slip tendency. These changes in parameter values all tend to decrease fault slip tendency relative to Case~1, though the results also depend on the particular random realization. Nonetheless, we do see $T_\text{s}$ values for Cases~2 and~3 that are lower than those for Case~1. Specifically, in Fig.~\ref{Fig: posterior slip tendencies for true models 2 and 3}, Case~2 is characterized by fault slip tendencies of 0.5--0.55, while those for Case~3 are even lower, 0.45--0.5. In both cases, DSI-VAE provides accurate predictions of fault slip tendencies, with fairly narrow ranges in the posterior histograms. This demonstrates the applicability of the DSI-VAE procedure to a variety of geological conditions.}

History matching results for the fault and geomechanical parameters for Cases~2 and~3 are provided in SI. Varying degrees of uncertainty reduction are achieved for these quantities. The general levels of posterior uncertainty are consistent with those for Case~1 shown in Fig.~\ref{Fig: posterior parameters for true model 1}.

\section{\textcolor{black}{Computational costs for DSI}}
\label{sec:costs}

\textcolor{black}{As noted in Section~\ref{Sec:DSI_results}, a full DSI run requires only about 7~minutes on a Nvidia A100 GPU. In contrast, a single GEOS run takes $\sim$2~hours on an AMD EPYC-7543 32-Core CPU. Our DSI-VAE workflow requires 1200~GEOS simulation runs and VAE training (which takes about 12~hours on a Nvidia A100 GPU). Thus the total computation for the workflow is about 2400~CPU hours plus 12~GPU hours. Because the GEOS runs can be performed in parallel, elapsed times are much less, with the speedup depending on the computational resources available.}

\textcolor{black}{One of the most efficient approaches for model-based history matching is to apply ESMDA directly to the geomodel. Typically, an ensemble size ranging from 200 to 1000 is used, and anywhere from 4 to 20 data assimilation steps are applied. These values correspond to total numbers of simulation runs ranging from 800 to 20,000. Thus, on the lower end, the computational effort is comparable to or even less than that for our workflow. Ensemble-based methods, however, entail Gaussian assumptions, and can provide inaccurate or nonphysical results when these assumptions are violated. In addition, numerical experimentation may be required to establish suitable localization strategies with ESMDA. \cite{teng2025likelihood}, for example, considered ESMDA with up to 200,000 simulation runs and demonstrated that their ESMDA implementation was unable to provide correct posteriors for systems with non-Gaussianity (as are considered here). More rigorous model-based methods such as Markov chain Monte Carlo, as used by, e.g., \citet{han2025accelerated}, do not assume Gaussianity and thus avoid these limitations. These approaches may require $\mathcal O(10^5)$ simulations, so they are much more computationally intensive than DSI-VAE or traditional ESMDA.}

\textcolor{black}{Our framework has the additional advantage that, once trained, it can be applied very quickly (7-minute run time) to assess the effect of different data types or measurement errors, the impact of additional or shifted monitoring wells, etc. With model-based history matching, each such assessment would require a full new run (e.g., $\mathcal O(10^3-10^4)$ GEOS simulations for ESMDA). Deep learning-based surrogate models are being increasingly used in place of high-fidelity simulators to accelerate history matching computations, though the construction of such a surrogate may be challenging and time-consuming for coupled problems of the type considered here. We note finally that a disadvantage of DSI is the lack of posterior geomodels, which may be required in some settings. Thus, the method of choice will depend on the types of decisions the workflow is intended to support.}

\section{Conclusions}\label{Sec: Conclusions}
In this work, we developed a DSI-VAE framework to predict pressure, strain, and stress fields in $\rm{CO}_2$ storage problems. This enables the computation of fault slip tendency, which is an important metric for predicting induced seismicity. Unlike traditional model-based history matching methods, the DSI framework directly generates posterior predictions for quantities of interest, thus significantly reducing computational costs. This makes the approach well-suited for computationally expensive coupled flow-geomechanics problems. 

The storage aquifer model in this study, which was based on a Gulf of Mexico geomodel, involved a 3D system with two extensive faults. The model consisted of a central storage aquifer, its surrounding formation, caprock, and basement rock. CO$_2$ was injected through three vertical wells, each injecting 1~Mt/year, for 50~years. Multiple heterogeneous permeability and porosity realizations, which included uncertain geomechanical and fault parameters, were constructed. Coupled flow-geomechanics simulations for a set of prior realizations, performed using GEOS, provided prior data for DSI and the VAE training data. The VAE was used to parameterize  pressure, strain ($zz$ component), and stress fields in terms of low-dimensional, near-Gaussian latent variables. Test-case results demonstrated that the VAE was able to reconstruct the target fields and generate new fields consistent with numerical simulation results.

The VAE representation was embedded into the DSI framework for history matching. Posterior (history matched) samples of the latent variables were generated using ESMDA. Pressure and strain data collected at four optimally located monitoring wells provided observation data. Our focus was on predicting pressure, strain, and stress fields at 50~years, along with fault slip tendency and key model parameters. 
Posterior results were generated for three cases involving different synthetic true models (additional results for Cases~2 and~3 are in SI). Our findings demonstrate that the DSI-VAE framework can provide accurate posterior predictions with significantly reduced uncertainty. Varying degrees of uncertainty reduction were observed for the fault and geomechanical parameters.

Several topics could be considered in future research. The framework should be applied to realistic models involving, for example, horizontal or deviated $\rm{CO}_2$ injection wells, a larger number of faults, and uncertain geological scenario parameters such as mean and variance of log-permeability. The monitoring well optimization strategy could be generalized, which might provide more uncertainty reduction for model parameters. Different types of observation data, such as $\rm{CO}_2$ saturation from monitoring wells, surface displacement measurements (for onshore projects), and seismic data, could be incorporated into the DSI-VAE methodology. This would allow us to evaluate the effect of different data types on posterior uncertainty, which is important in the design of monitoring strategies.

\section*{Acknowledgments}

We are grateful to the Stanford Center for Carbon Storage and the Stanford Smart Fields Consortium for financial support. We thank Oleg Volkov, Jian Huang and Oluwatobi Raji for assistance with the geomodel and GEOS, and Wenchao Teng, Nanzhe Wang, and Yifu Han for useful discussions. We acknowledge the SDSS Center for Computation for providing HPC resources.

\section*{Code availability}

The code used in this study is available at: \url{https://github.com/xiaowen012/DSI_VAE}.

\section*{CRediT authorship contribution statement}
{\bf Xiaowen He}: Writing – original draft, Visualization, Software, Methodology, Investigation, Formal analysis, Data curation, Conceptualization.
{\bf Su Jiang}: Writing – review \& editing, Software, Methodology, Conceptualization.
{\bf Louis J. Durlofsky}: Writing – review \& editing, Supervision, Resources, Funding acquisition, Formal analysis, Conceptualization.

\bibliographystyle{cas-model2-names}
\bibliography{bibliography} 

\end{document}